# Object Detection in Specific Traffic Scenes using YOLOv2

Shouyu Wang, Weitao Tang

*Abstract*—Real-time object detection framework plays crucial role in autonomous driving. In this paper, we introduce the real-time object detection framework called You Only Look Once (YOLOv1) and the related improvements of YOLOv2. We further explore the capability of YOLOv2 by implementing its pre-trained model to do the object detecting tasks in some specific traffic scenes. The four artificially designed traffic scenes include single-car, single-person, frontperson-rearcar and frontcar-rearperson. YOLOv2 model performs well in detecting single-car once scope of camera distance is within 60 feet. For detecting single person, YOLOv2 handle most of situations when camera distance is within 50 feet. For detecting both person and car, YOLOv2 always detect unsuccessfully if person overlaps with car in photo. In frontcar-rearperson scenario, the "success rate of prediction" goes down sharply in 50 feet camera distance which is shorter than the same situation in frontperson-rearcar scenario.

## I. INTRODUCTION

Object detection is one of the most popular research topics in computer vision field. Amount of object detection techniques has been developed in the last couple years. There are various applications of these techniques, such as robotic navigation [1], video stream analysis [2], autonomous driving [3], and structural visual inspection [4]. Real-time object detection framework plays crucial role within scope of the fundamental algorithms in autonomous driving field as well [3]. It's an efficient method to percept surrounding vehicles, passengers, traffic lights or other unknown objects. Recently based on widely developments and implements of deep learning technique, object detection related frameworks have achieved significant improvements.

Throughout the progress of hardware development and deep learning theories, the barrier of developing and implementing CNN (convolutional neural network) based frameworks for solving physical problems has been going down gradually. For now, CNN-based methods for object detection have achieved significant improvements. There are several varieties of CNN based object detector for choosing, such as R-CNN [5], Fast R-CNN [6], Faster R-CNN [7], Light-Head R-CNN [8], Cascade R-CNN [9], SPP-Net [10], MR-CNN [11], YOLO [12], YOLOv2 [13], YOLOv3 [14], SSD [15], DSSD [16], R-FCN [17], MS-CNN [18], etc.

CNN-based methods can be commonly classified into two categories, namely, one-stage framework and two-stage framework. The most obvious feature for differentiating them is whether or not the framework trained by end-to-end. Most of time one stage framework can process end-to-end training and two stages framework cannot.

Generally two stage framework, such as R-CNN [5], Fast R-CNN [6], Faster R-CNN [7], Light-Head R-CNN [8], Cascade R-CNN [9], R-FCN [17], MS-CNN [18], would generate several proposal regions firstly. Then the framework would concentrate on classifying the specific categories of proposal regions. Single-stage frameworks usually combines the processes mentioned above into an end-to-end training framework such as YOLO [12], YOLOv2 [13], YOLOv3 [14], SSD [15], DSSD [16]. Generally single stage frameworks over-perform two stages frameworks in terms of detecting speed.

The main criterions for judging the performance of object detector cover accuracy and speed. Most of time different state-of-the-art frameworks are trying to utilize bunch of techniques for achieving best trade-off between these two aspects. The tasks of object detection are complicated as the detector solves both classification and localization tasks. It is hard to achieve both goals simultaneously.

Object detection is a crucial task in autonomous driving field. Generally valid detection results in reliable information for further inference and utilization. For autonomous driving there are also some basic requirements for these frameworks such as accuracy, fast, small. Accuracy, ideally we would like the detector to achieve high precision on objects of interests. For fast, it means that it should be able to do inference in real-time. It's good for reducing latency of vehicle control loop as well. For size, the detector should be small so that it could be more accessible for embedded system deployment. It will also trigger just less energy consumption. For the sake of having advantages of single stage framework and satisfying requirements mentioned above, YOLO series of models have become super popular object detection methods implementing in autonomous driving field.

Girshick et al. combined region proposals with CNNs and proposed an object detection method called R-CNN (regions with CNN features) [5]. R-CNN use deep learning approaches to identify regions that may contain objects. R-CNN requires computing each region proposal and thus, it is inefficient. Even though there are various updated R-CNN versions (e.g., SPP-Net [10], Fast R-CNN [6], Faster R-CNN

S. Wang is with the School of Engineering Technology, Purdue University, West Lafayette, IN 47907 USA (e-mail: wang3765@purdue.edu).

W. Tang is with the School of Computer and Information Technology, Purdue University, West Lafayette, IN 47907 USA (e-mail: tang384@purdue.edu).

[7]) that have been developed to improve the efficiency of R-CNN, R-CNN is still not efficient enough for real-time object detection.

Redmon et al. developed an object detection model called You Only Look Once (YOLO) [12]. YOLO integrates region proposition and classification into a single stage, which speeds up the model drastically and makes it applicable for real-time object detection. Improved versions of YOLO (e.g., YOLOv2 and YOLO9000 [13]) have also been released. YOLOv2 has been implemented in Apollo repository, an open source autonomous driving platform. The series of YOLO frameworks advanced the research on autonomous driving. Jensen et al. had explored how to use YOLO models to detect traffic light [19]. Usage of YOLO models in other traffic scenes has not been reported in the literature.

In this paper, we investigate the performance of the state-of-the-art single stage YOLOv2 real-time object detection framework in various traffic scenes. More specifically, we design four different traffic scenes: single-person , single-car, front-person-rear-car, and front-car-rear-person. There are four different experiments have done in judging the performance of YOLOv2 model in these traffic. Based on the results of experiments, we analysis the characteristic of YOLOv2 in detecting different kinds of objects. Then comparing with result of experiment 3, some further explorations are also done in figuring out the cause of earlier "sharp dropping" in experiment 4. In final, the direction of further improving of YOLO series of frameworks are also being more clear base on the research we have made.

The rest of the paper is organized as follows. Section II introduces the methodology of the YOLO and YOLOv2 object detection frameworks. Section III presents the details of four different specific traffic scene experiments. Experimental results are analyzed and discussed in section IV. Finally, the paper is summarized in section V.

## II. METHODOLOGY

A YOLO object detection model serves an entire image as its input. The outputs of model will be the coordinates of the bounding box and the classification of the bounding box. Unlike R-CNN that processes the proposal and classification separately, YOLO model processes the localization, detection, and classification simultaneously. Hence, YOLO model is capable of processing end-to-end training that has simpler structure and consumes less space in storing related data.

### A. How a YOLO Model Works

Once we utilize YOLO model to do the object detection, the input image will be divided into $S \times S$ grid cells, where $S$ is an integer. Suppose the center of a detected object is located in one of the grid cells, then the classification of the object is determined by the specific grid cell.

Each grid cell will be in charge of predicting B bounding boxes. After calculating, YOLO model completes the regression of each bounding box. Each bounding box should predict a value of confidence as well. The confidence value indicates whether or not there is an object in the bounding box and how accurate the prediction is. The value calculated by $Pr(Object) \times IOU_{pred}^{truth}$. $Pr(Object)$ is the possibility that object exist in specific grid cell. When there is an object located in grid cell, $Pr(Object) = 1$. Otherwise, $Pr(Object) = 0$. $IOU_{pred}^{truth}$ indicates the similarity between the appropriated bounding box and the ground-truth bounding box. In conclusion, each bounding box predicts values of five parameters, namely, $x, y, w, h$, and confidence.

Each grid cell predicts the classification the object. If there are $m$ categories of objects, the classification of an object needs to take $m$ data. Therefore, to predict $n$ bounding boxes and $m$ categories for an image with $S \times S$ grid cells, the output of the YOLO model should be a $S \times S \times (5 \times n + m)$ tensor.

We can get the category-specific confidence score of each bounding box by multiplying the category value and the confidence value in the detecting process. The process can be explained as

$$Pr(Class_i|Object) \times Pr(Object) \times IOU_{pred}^{truth}$$
$$= Pr(Class_i) \times IOU_{pred}^{truth}. \quad (1)$$

where $Pr(Class_i|Object)$ is the prediction of categories. The $Pr(Object) \times IOU$ is the confidence of each bounding box. Once the category-specific confidence score is calculated, one can set up a threshold to filter the bounding boxes with low scores. Then Non-max Suppression (NMS) can be utilized to process the remaining bounding boxes and get the final prediction results.

### B. Training protocol

In YOLO, the data of each grid cell would be stored in a 30 dimensions vector. Specifically, eight dimensions represent the coordinates of bounding box, 2 dimensions represent the value of confidence, and the remaining 20 dimensions represent the probability of different categories. Moreover, the coordinate *x*, *y* are normalized into a range from 0 to 1 according to related offset value. The value of *w, h* should do the normalization into a range from 0 to 1 according to the width and height of image.

In training protocol, constructing a loss function that can balance the training of predicting related coordinate, confidence and category is very important. In the very first time, the designer construct the loss function by concatenating three sum-squared error parts. This method will also bring some problems. First, the weight of localization error which has 8 dimensions and the weight of classification error which has 20 dimensions shouldn't be the same. Then, generally most grid-cell is without any objects so that it is easy to result in unstable results or divergence. For solving these problems, the designer set larger loss weight to the term of coordinate prediction that gets the value of $\lambda_{coord}$ as 5. Meanwhile they reduced the loss weight of boxes that is without objects by setting the value of $\lambda_{noobj}$ as 0.5. For the boxes that have objects and its loss, the designer sets the loss weight as 1.

Another problem caused by the bias of predicted box. The influence of specific bias to large box is different from the same bias to tiny box totally. However, the same bias will cause same variance of loss in sum-square error loss function. For solving this, designer use the square root of width and height to calculate the loss function.

Finally, the entire loss function can be expressed as

$$Loss = \lambda_{coord} \sum_{i=0}^{s^2} \sum_{j=0}^{n} I_{ij}^{obj} (x_i - \hat{x}_i)^2 + (y_i - \hat{y}_i)^2$$

$$+\lambda_{coord} \sum_{i=0}^{s^2} \sum_{j=0}^{n} I_{ij}^{obj} \left(\sqrt{w_i} - \sqrt{\widehat{w}_i}\right)^2 + \left(\sqrt{h_i} - \sqrt{\hat{h}_i}\right)^2$$

$$+\sum_{i=0}^{s^2} \sum_{j=0}^{n} I_{ij}^{obj} (m_i - \widehat{m}_i)^2 + \lambda_{noobj} \sum_{i=0}^{s^2} \sum_{j=0}^{n} I_{ij}^{noobj} (m_i - \widehat{m}_i)^2$$

$$+\sum_{i=0}^{s^2} I_{ij}^{obj} \sum_{m \in classes} \left(p_i(m) - \hat{p}_i(m)\right)^2 \quad (2)$$

Where $\lambda_{coord}$ is the weight of the loss of the coordinates of a boundary box. $\lambda_{noobj}$ is the weight of the loss when detecting background. $I_{ij}^{obj}$ is the $j$th bounding box predictor that is in the $i$th cell is valid in prediction. $s^2$ is the number of grid cells. $B$ is the number of bounding boxes to be predicted in each grid cell. $x_i$ and $y_i$ denote the coordinates of the base point of the $i$th bounding box. $w_i$ and $h_i$ denote the weight and height of the $i$th bounding box. $C_i$ is the confidence score of the $j$th bounding box. $p_i(c)$ is the conditional class probability for category $c$.

The loss function has three main parts, namely, coordinate prediction, confidence prediction, and classification prediction. These three predictions work only if an object is in the bounding box. In a grid cell, only the bounding box that has highest *IOU* is in charge of the ground truth bounding box and process learning from coordinate error.

### C. Improvments of YOLOv2

YOLOv2 made improvements over YOLO to overcome the following two shortcomings: 1) localization precision of object still full of space of improving, 2) recall is lower than the framework that based on region proposal.

In term of model complexity, YOLOv2 is actually simpler than YOLO. YOLOv2 has also made many improvements, such as batch normalization, high resolution classifier, convolutional with anchor boxes, dimension clusters, direct location prediction, fine-grained features, and multi-scale training. Base on the training in VOC2007, with these improvements, the performance of YOLOv2 has been improved to 78.6 *mAP*. Moreover, the designer of YOLOv2 proposed a new backbone called Darknet-19 which has fewer parameters than GoogleNet [20] which is used in YOLO. YOLOv2 also has a new robust mechanism that allows it be trained by detection data and classification data simultaneously.

## III. EXPERIMENTS

Capabilities of detecting vehicles and pedestrians are crucial for frameworks (e.g., YOLOv2) utilized in autonomous driving. In this section, we focus on the understanding of how the pre-trained YOLOv2 model performs in traffic scenes. The official pre-trained YOLOv2 model is used to detect targets (e.g., a person or a car) in various traffic scenes.

### A. Pre-Trained Model

The YOLOv2 model pre-trained using the COCO dataset can be accessed from the official website of YOLOv2. This pre-trained YOLOv2 model can get 76.8 *mAP* at 67 FPS or 78.6 *mAP* at 40 FPS on VOC 2007 [13]. Referring to the configuration and weight files of the official pre-trained YOLOv2 model, we create a YOLOv2 model in our study using Pytorch and utilize this outperforming to process the further specific detect tasks.

### B. Experiments in Specific Traffic Scenes

We investigates the performance of the YOLOv2 model in detecting humans and cars with experiments. How the relative locations of humans and cars affect the performance of the YOLOv2 model is also investigated. Four experiments are designed. They are conducted to detect a single person in experiment 1, a single car in experiment 2, a person behind a car in experiment 3, and a car behind a person in experiment 4, respectively.

We define "camera distance" as the distance from camera to the baseline in the further experiments we make. In experiments 1 and 2, each person or car has three positions: left, middle, and right. A camera is used to take photos of the person or the car. The camera distance is 10, 20, 30, 40, 50, 60 feet respectively (shown in figure 1 and figure 2). In experiment 3, the car will move along the base-line separately which covers 3 positions. The person moves separately in the position matrix which covers 9 different positions. In experiment 4, the person will move separately along the base-line which covers 3 points. The car moves in the position matrix separately which only has 6 positions (without 3 positions that will block the person when taking the photo with camera). A camera is used to take photos with the person and the car simultaneously in experiments 3 and 4. The camera is mounted in 40, 50, 60 feet camera distance respectively (shown in figure 3 and figure 4). In each specific position pair of the car, the person, and the camera, three photos will be taken. Finally, we take 507 photos in total and they are both utilized for doing the object detecting tests of YOLOv2 model. The distance between adjacent positions in the position matrix is 10 feet.

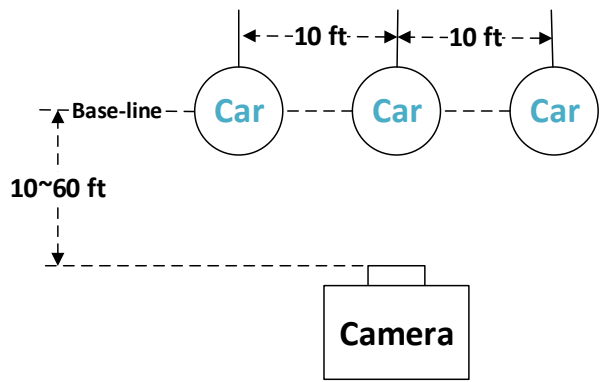

Figure 1. Illustration of experiment 1

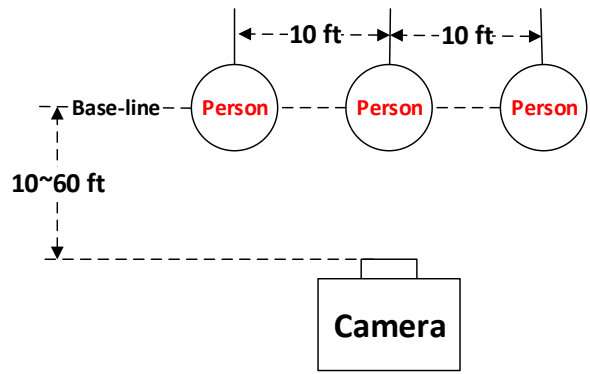

Figure 2. Illustration of experiment 2

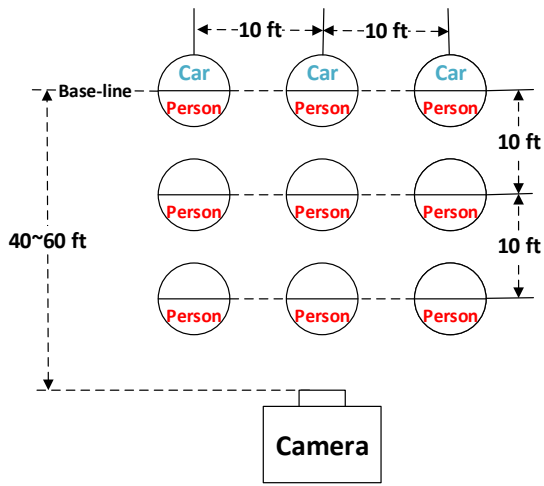

Figure 3. Illustration of experiment 3

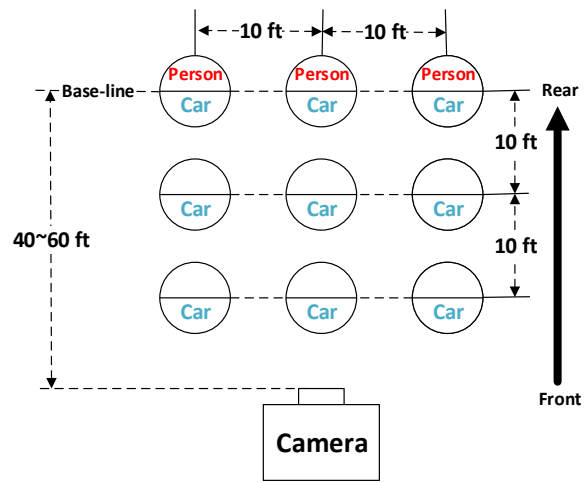

Figure 4. Illustration of experiment 4

## IV. RESULTS

In each specific position pair of person and car, we will shoot 3 differ photos for further detecting. In chart of experiment result of 1 and 2, we list all the detection results in each specific position pair. In chart of experiment result of 3 and 4, we only list the "final result" in each specific position pair. For each specific position pair, we only record the result as "success" (shown as √ in chart) when YOLOv2 detects all the targets (including person and car) successfully. The result which is in most in the same specific position pair will be served as the "final result". Final result will be used for further calculating the "success rate of prediction". The "Camera Distance" in the chart means the distance between camera and base-line.

In detection result chart of experiment 1, YOLOv2 performs greatly in detecting a single person when the camera distance is in 40 feet. YOLOv2 can still handle part of situations in 50 feet camera distance. When the camera distance reaches 60 feet, YOLOv2 performs poorly in detecting person.

In detection result chart of experiment 2, YOLOv2 performs greatly in detecting a car when the scope of camera distance is within 60 feet. There is only one "failure" (shown as × in chart) in 60 feet camera distance.

In detection result chart of experiment 3, YOLOv2 performs well in detecting the car in all camera distances. YOLOv2 can detect the person as well when the camera is within 50 feet camera distance. However, once person overlaps with car, regardless they are in the left, middle or right, YOLOv2 gets "failure" in final.

In the results of experiment 4, YOLOv2 shows stable performance when the camera distance is 40 feet. Once the camera distance goes up to 50 feet, the detecting begins to emerge some unstable results. When the camera distance goes to 60 feet, YOLOv2 almost fully lost capability in detecting car and person simultaneously.

Base on comparing between experiment 3 and 4, we see the position of person and car do affect the detection result. As seen in figure 6, in frontperson-rearcar scenario, YOLOv2 gets success rate of 81.48% in 40 feet camera distance. The success rate is still at 77.78% when the camera distance comes to 50 ft. The success rate sharply goes down to 44.4% when the camera distance comes to 60ft. This means that 60 feet is really a far distance for YOLOv2 to do the detecting for both person and car simultaneously in this scenario. In scenario of frontcar-rearperson, YOLOv2 get excellent performance in 40 feet camera distance with 100% success rate. Nevertheless, it only get 33.33% success rate in 50 feet camera distance. It means that, in this scenario, YOLOv2 can only handle detecting task within scope of 40 feet camera distance.

The causes that affect the success rate of YOLOv2 deserve further discussion. As we can see in previous experiment 1, YOLOv2 shows excellent performance in detecting the car within scope of 60 feet camera distance. In experiment 2, YOLOv2 can only handle the scope of 50ft camera distance in most of times. It shows that YOLOv2 is better in detecting bigger object in same camera distance. In experiment 3 and 4, the successful detection of car will not be the key point to improve success rate as YOLOv2 already shows great capability in detecting car within 60 feet camera distance in experiment 1. The key point for improving success rate is depending on the successful detection of person which is small in photo. In experiment 4, the size of person will be smaller than the size in experiment 3 so that "sharp dropping" of success rate will happens in shorter camera distance. With the influence of bigger size of car in picture, YOLOv2 can only get poorly 33.33% success rate at 50 feet camera distance in experiment 4.

TABLE II. DETECTION RESULTS OF EXPERIMENT 3

| Experiment 3 | | Camera Distance (feet) | | | | | | | | |
|---|---|---|---|---|---|---|---|---|---|---|
| | | Left Car | | | Middle Car | | | Right Car | | |
| | | 40 | 50 | 60 | 40 | 50 | 60 | 40 | 50 | 60 |
| Person Position (feet) | | | | | | | | | | |
| 0 | L | × | × | × | √ | √ | √ | √ | √ | × |
| 0 | M | √ | √ | √ | × | × | × | √ | √ | × |
| 0 | R | √ | √ | × | √ | × | √ | × | × | × |
| 10 | L | √ | √ | × | √ | √ | √ | √ | √ | × |
| 10 | M | √ | √ | × | × | × | × | √ | √ | √ |
| 10 | R | √ | √ | × | √ | √ | × | √ | √ | × |
| 20 | L | √ | √ | √ | √ | √ | × | √ | √ | √ |
| 20 | M | √ | √ | √ | × | × | × | √ | √ | √ |
| 20 | R | √ | √ | √ | √ | √ | √ | √ | √ | √ |

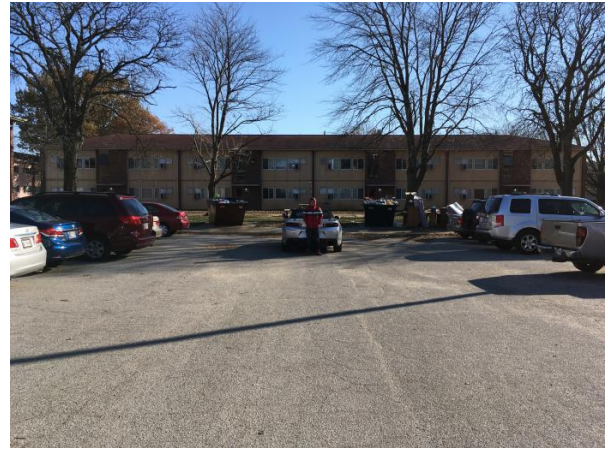

Figure 5. The situation that person and car are overlapping

TABLE I. DETECTION RESULTS OF EXPERIMENTS 1 AND 2

| Experiment 1 | | Camera Distance (feet) | | | | | |
|---|---|---|---|---|---|---|---|
| | | 10 | 20 | 30 | 40 | 50 | 60 |
| Single Person | Left | — | √√× | √√√ | √√√ | √√√ | ××× |
| | Middle | √√√ | √√√ | √√√ | √√√ | √×× | ××× |
| | Right | — | √√√ | √√√ | √√√ | √√× | √√√ |

| Experiment 2 | | Camera Distance (feet) | | | | | |
|---|---|---|---|---|---|---|---|
| | | 10 | 20 | 30 | 40 | 50 | 60 |
| Single Car | Left | √√√ | √√√ | √√√ | √√√ | √√√ | √√√ |
| | Middle | √√√ | √√√ | √√√ | √√√ | √√√ | ×√√ |
| | right | √√√ | √√√ | √√√ | √√√ | √√√ | √√√ |

TABLE III. DETECTION RESULTS OF EXPERIMENT 4-1

| Experiment 4-1 (left person) | | Camera Distance (feet) | | |
|---|---|---|---|---|
| | | 40 | 50 | 60 |
| Car Position (feet) | | | | |
| 0 | M | √ | × | × |
| 0 | R | √ | × | √ |
| 10 | M | √ | × | × |
| 10 | R | √ | √ | × |
| 20 | M | √ | × | × |
| 20 | R | √ | √ | × |

TABLE IV. DETECTION RESULTS OF EXPERIMENT 4-2

| Experiment 4-2 (middle person) | Camera Distance (feet) | | |
|---|---|---|---|
| | 40 | 50 | 60 |

| Car Position (feet) | 0 | L | ✓ | × | × |
|---|---|---|---|---|---|
| | | R | ✓ | × | × |
| | 10 | L | ✓ | ✓ | × |
| | | R | ✓ | × | × |
| | 20 | L | ✓ | ✓ | × |
| | | R | ✓ | ✓ | × |

TABLE V. DETECTION RESULTS OF EXPERIMENT 4-3

| Experiment 4-3 (right person) | | | Camera Distance (feet) | | |
|---|---|---|---|---|---|
| | | | 40 | 50 | 60 |
| Car Position (feet) | 0 | M | × | × | × |
| | | L | ✓ | × | × |
| | 10 | M | ✓ | × | × |
| | | L | ✓ | × | × |
| | 20 | M | ✓ | ✓ | ✓ |
| | | L | ✓ | ✓ | ✓ |

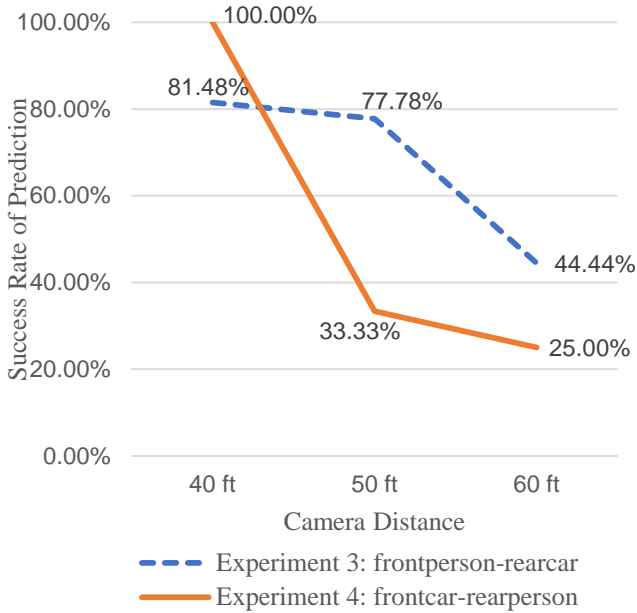

Figure 6. Comparison of prediction success rate of experiments 3 and 4

## V. CONCLUSION

In this paper, we introduced the state-of-the-art object detection models: YOLO and YOLOv2. Furthermore, we explained how a YOLO model divides an image into multiple grid cells and how these grid cells process the objectness, localization, and classification. We also introduced the training protocol and techniques of constructing loss function. There is also some further explanations about how YOLOv2 do the simplification to its model and improvement in preprocessing, training, and objection detection.

In experiments, based on pre-trained YOLOv2 model, we do the object detection in specific traffic scenes. The traffic scenes include single cars, single person and specific pair of person and car. Experiment results showed that YOLOv2 gets excellent capability in detecting single cars within the scope of 60 feet camera distance. For detecting single person, YOLOv2 can also achieve great results within scope of 50 feet camera distance. The situations of doing object detection about specific pair of car and person will be more sophisticated. When the car overlaps with person in the photo, YOLOv2 will always performs failure no matter where the pair of person and car is. YOLOv2 is good at detecting big object like car in photo and not so good at detecting smaller object like person. It's also the reason why in experiment 4, the "sharp dropping" of success rate of prediction happens in only 50 feet camera distance which is shorter than the situation in experiment 3.

Based on previous experiments we have been discussed with, we already touch the boundary of capability of YOLOv2 in doing object detection in some specific traffic scenes. Further improvement could be made in the direction of how to further push the capability of YOLO series of frameworks forward in detecting smaller objects in photo.